\relax
\documentclass[letterpaper]{article} % DO NOT CHANGE THIS
\usepackage{aaai21}  % DO NOT CHANGE THIS
\usepackage{times}  % DO NOT CHANGE THIS
\usepackage{helvet} % DO NOT CHANGE THIS
\usepackage{courier}  % DO NOT CHANGE THIS
\usepackage[hyphens]{url}  % DO NOT CHANGE THIS
\usepackage{graphicx} % DO NOT CHANGE THIS
\usepackage{natbib}  % DO NOT CHANGE THIS AND DO NOT ADD ANY OPTIONS TO IT
\usepackage{caption} % DO NOT CHANGE THIS AND DO NOT ADD ANY OPTIONS TO IT
\usepackage{array, amsmath}
\usepackage{tabularx,booktabs}

\usepackage[switch]{lineno}

\newcolumntype{P}[1]{>{\centering\arraybackslash}p{#1}}
\newcolumntype{R}{>{\raggedleft\arraybackslash}X}

\title{Binding and Perspective Taking as Inference\\
	in a Generative Neural Network Model}
\author{
    Mahdi Sadeghi\textsuperscript{\rm 1}, Fabian Schrodt\textsuperscript{\rm 1}, 
    Sebastian Otte\textsuperscript{\rm 1}, Martin V. Butz\textsuperscript{\rm 1}
    \\
}
\affiliations{
	\textsuperscript{\rm 1}University of T{\"u}bingen\\
	Neuro-Cognitive Modeling Group\\
	Sand 14, D-72076, T{\"u}bingen, Germany\\
	\{mahdi.sadeghi$\mid$tobias-fabian.schrodt$\mid$sebastian.otte$\mid$martin.butz\}@uni-tuebingen.de

}
\begin{document}

%\linenumbers

\maketitle

\begin{abstract}
The ability to flexibly bind features into coherent wholes from different perspectives is a hallmark of cognition and intelligence.
Importantly, the binding problem is not only relevant for vision but also for general intelligence, sensorimotor integration, event processing, and language.
Various artificial neural network models have tackled this problem with dynamic neural fields and related approaches.
Here we focus on a generative encoder-decoder architecture that adapts its perspective and binds features by means of retrospective inference.
We first train a model to learn sufficiently accurate generative models of dynamic biological motion or other harmonic motion patterns, such as a pendulum.
We then scramble the input to a certain extent, possibly vary the perspective onto it, and propagate the prediction error back onto a binding matrix, that is, hidden neural states that determine feature binding.
Moreover, we propagate the error further back onto perspective taking neurons, which rotate and translate the input features onto a known frame of reference.
Evaluations show that the resulting gradient-based inference process solves the perspective taking and binding problem for known biological motion patterns, essentially yielding a Gestalt perception mechanism.
In addition, redundant feature properties and population encodings are shown to be highly useful.
While we evaluate the algorithm on biological motion patterns, the principled approach should be applicable to binding and Gestalt perception problems in other domains.
\end{abstract}

\section{Introduction}
Social cognition depends on our ability to understand the actions of others.
The simulation theory of social cognition \cite{Barsalou:1999,Gallese:1998,Johnson:2005} suggests that visual and other sensory dynamics are mapped onto the own sensorimotor system.
The mirror neuron system has been proposed to play a fundamental role in this respect.
It appears to project and interpret visual information about others with the help of ones own motor repertoire \cite{gallese2004unifying}.
\citet{Cook:2014} argue that the involved mirror neurons can develop purely via associative learning mechanisms.
Two fundamental challenges remain, however, when attempting to learn such associations: 
the perspective taking problem and the binding problem.

The \emph{perspective taking problem} addresses the challenge that visual information about others comes in a different frame of reference than information about ones own body.
It has been shown that we have a strong tendency to adapt the perspective of others, particularly when observing interactions with other entities \cite{Tversky:2009}.
These and related results and modeling work suggest that our brain is able to project our own perspective into the other person by some form of transformation \cite{Johnson:2005,Meltzoff:2002,schrodt2014modeling}. 
Indeed, in adults this ability it not restricted to spatial transformations but extends to the adoption of, for example, conceptual and affective standpoints \cite{Moll:2011}.
When focusing on spatial perspective taking, the challenge is to transform an observed action into a canonical perspective, to thus prime corresponding motor simulations \cite{castiello2002observing,edwards2003motor}.

Besides the perspective taking challenge, the \emph{binding problem} \cite{Butz:2017,treisman1998feature} poses another severe challenge.
In particular, individual visual features, such as motion patterns, color, texture, or edge detectors, 
have to be integrated into a complete Gestalt \cite{Jaekel:2016,koffka2013principles} in order to recognize an entity, such as the body of another person.
With respect to biological motion recognition, the problem has also been termed the \emph{correspondence problem} \cite{nehaniv2002correspondence,heyes2001causes}: Which observed body part corresponds to which own body part?  
Action understanding and imitation appears to be facilitated by establishing correspondences between the own body schema and the one of an observed person \cite{jackson2006neural}.
 	% fMRI studies have suggested that neurons in the lateral occipitotemporal cortex are specifically tuned to process information about the human body \cite{downing2001cortical}.
To bind the observed features together into an integrated bodily percept, top-down expectations are interacting with bottom-up saliency cues in an approximate Bayesian manner \cite{buschman2007top,Jaekel:2016}. 
The details of the involved processes, however, remain elusive. 

%5mechanisms related to interactive attention appear to be at workted to attention and are driven by top-down task demands and bottom-up saliency cues 
% The visual system leads attention to particular salient features and binds them in a manner that important scene information are recognized. 
%In conjunction with mirror neurons, ,Milner:2008}.

Here we propose a generative, autoencoder-based neural network model, which solves the perspective taking and binding problems concurrently by means of retrospective, prediction error-minimizing inference.
The encoder part of the model is endowed with a transition vector and a rotation matrix for perspective taking and a binding matrix for flexibly integrating input features into one Gestalt percept. 
The parameters of these three modules are tuned  online by means of retrospective inference \cite{Butz:2019}.
As a result, the model mimics approximate top-down inference, attempting to integrate all bottom-up visual cues into a Gestalt from a canonical perspective. 
The model is first trained on a canonical perspective of an ordered set of motion features, simulating a self-grounded Gestalt perception \cite{gallese2004unifying}.
Next, perspective taking and feature binding is evaluated by projecting the reconstruction error back onto the perspective and binding parameters, which can be viewed as specialized parametric bias neurons \cite{Sugita:2011a,Tani:2017}.
Our evaluations show that it is highly useful (i) to split the motion feature information into relative position, motion direction, and motion magnitudes and (ii) to use population encodings of the individual features. 
We evaluate the model's abilities on a two dimensional, two joint pendulum and on three dimensional cyclic dynamical motion patterns of a walking person.

%%%%  THE MODEL %%%%%%
\section{Proposed Model}
The proposed architecture has some fixed components and some other components that are trained or adapted online by backpropagation.
The model learns a generative, autoencoder-based model of motion patterns, which it employs to bind visual features and perform spatial perspective taking.
% It processes a number of observed salient features.
In our case, each visual feature corresponds to a joint location and is represented by a Cartesian coordinate relative to a global six-dimensional frame of reference (encoding origin and orientation). 
Figure~\ref{fig1:model} shows a sketch of the processing and inference architecture.\footnote{A previous version of the specified architecture, which implemented a different autoencoder, is available electronically as a PhD thesis \cite{Schrodt:2018}. The model has never been submitted to a conference or journal before.} 

\subsection{Sub-Modal Population Encoding}
\begin{figure}[t!]
\centering
\includegraphics[width=0.98\columnwidth]{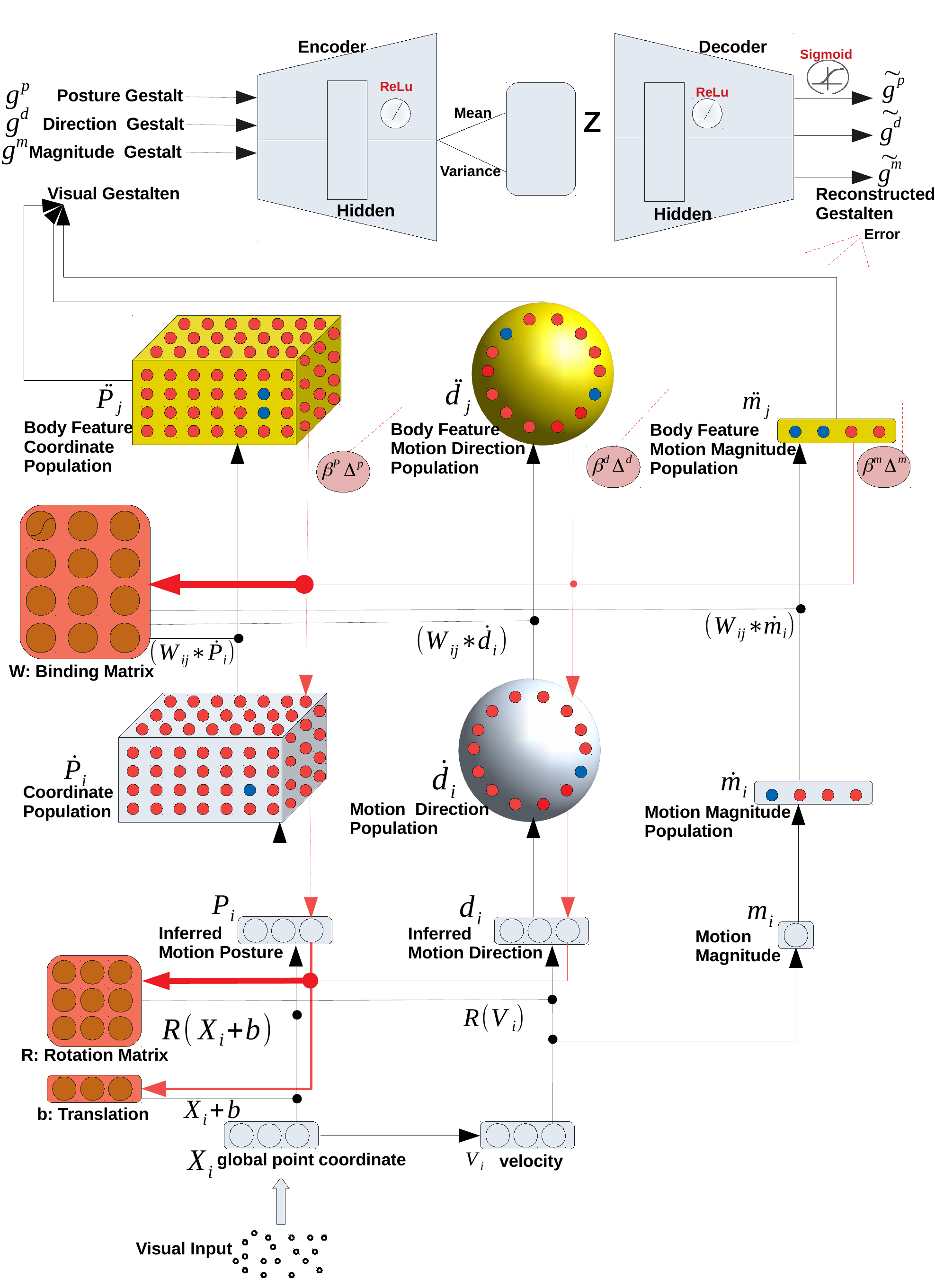}
\caption{
Our generative, modularized neural network model processes a visual 
	feature $i$ by first applying possible translation and rotation operations. 
	Next, the feature is encoded into respective population codes. 
	A neural gating matrix binds feature $i$ to an autoencoder input slot $j$.
	Variational autoencoders for posture, motion direction, and magnitude  attempt to reconstruct the observed patterns. 
	The squared difference between input and reconstruction is used as the loss signal to adjust the parameters of the gating and rotation matrices and the translation vector via retrospective inference.
}
%Illustration of the proposed model in a connectivity graph that processes a single visual feature, 
%At first the i-th input feature will be transformed with rotation and translation, then its corresponding inferred posture, motion direction, and motion magnitude submodalities are extracted and then encoded indivudually by arranged population of neurons. Afterwards, a binding matrix, that is, hidden neural states which determine feature binding assigns the observed input feature to the correct population, that is, j-th bodily information. 
%Translation Rotation and Binding matrix are adapted by top-down error signals that stem from self-generated submodal expectations}
\label{fig1:model}
\end{figure}

Each Cartesian input coordinate at each time step $t$ is 
% MVB: the following is too much detail: 
% 	represented by activity of a layer consisting of three neurons (one neuron per dimension) and is 
segregated into three distinct sub-modalities (i.e. posture, motion direction, and motion magnitude). 
% MVB: let's keep the order posture, direction magnitude! (throughout the paper)
% MVB: Velocity should be obvious, no need for the equation!
%The network calculates the velocity $V_i$ of each visual feature coordinate accoding to the following formula:\\
%$V_{t} = X_{t} - X_{t-1}$
%\begin{equation}
%V_{t} = X_{t} - X_{t-1}
%\end{equation}
Note that the relative position of a visual feature depends on the choice of origin and the orientation of the coordinate frame.
Moreover, while motion direction depends only on the orientation but not on the origin, motion magnitude is completely independent of perspective. 
% The model will learn partially invariant codes from these distinct submodal perceptions and use them for perceptual inference.
As a result, given a visual input coordinate and its velocity, three types of submodal information are derived
each transformed via a rotation matrix $R$ and a translation bias $b$, which determine orientation and origin, that is, the frame of reference, respectively. 
%, both of which are applied to the whole visual percept, that is, all features consistently. The rotation matrix models the capability to imagine different perspectives or vantage points on observed visual features, while the translation bias represents the center of the respective coordinate system, which is also the center of rotation. 
At each point in time $t$, the transformations are applied as follows:
\begin{equation}
P_{i}(t) = R(t) \cdot X_{i}(t) + b(t),
\end{equation}
determining the relative position $P_i(t)$ of the feature $X_{i}(t)$;
\begin{equation}
m_{i}(t) =\left \| R(t) \cdot V_i(t) \right \|,
\end{equation}
where $V_{t} = X_{t} - X_{t-1}$ denotes the velocity to determine the absolute motion magnitude $m_i(t)$;
\begin{equation}
d_{i}(t) = \frac{R(t) \cdot V_{i}(t)}{m_{i}(t)},
\end{equation}
calculating the relative motion direction $d_i(t)$.
Figure~\ref{fig1:model} shows a sketch of the proposed model in a connectivity graph, including the processing pipeline for a single three dimensional visual feature.

After the extraction of submodal information and the projection onto a particular visual frame of reference, each submodality is encoded separately by one population of topological neurons with Gaussian tunings, which yield local responses to specific stimuli within a limited range.
Such encodings can be closely related to encodings found in the visual cortex and beyond \cite{Olshausen:1997,pouget2000information}.
%After the extraction of the sub-modal information and inference of a visual frame of reference, each sub-modality is encoded separately by one population of topological neurons with Gaussian tunings, which represent local receptive fields to specific stimuli within a limited range. 
However, to assess the effectiveness of population coding, we also evaluate the model's performance on the raw submodal information.
The topological neurons of posture, direction, and magnitude populations are evenly projected onto a cube, a unit sphere, and a line, respectively, such that the input space is fully covered. 
% MVB: Careful with "optimally" - we never checked is this is "optimal": 
%          in a way that they cover the input space optimally.
According to the motion capture data and the applied skeleton, posture, direction, and magnitude populations are configured to contain $64$, $32$, and $4$ neurons, respectively.
In the case of the two jointed pendulum experiments, we distributed $16$ posture neurons on a rectangle area around the stimuli, $8$ direction neurons on a unique circle, and $4$ magnitude neurons.
The response $\smash{\dot{P}_{\alpha i}}$ of the $\alpha$-th neuron in a population that encodes the position $p$
of the i-th visually observed feature is computed by:
\begin{equation}
\dot{P}_{\alpha,i}(t)=(r^{p})^{D^{p}} \cdot N\left(P_{i}(t); c_{\alpha}^{p}, \Sigma^{p}\right)
\end{equation}
where $N(l; \mu, \Sigma)$ is the density of the multivariate normal distribution at $l$ with mean $\mu$ and a $D^{p}$-dimensional diagonal covariance matrix $\Sigma$. 
Each neuron in a submodal feature population has an individual center $c_{\alpha}^{p}$ as well as a response variance $\Sigma^{p}$.
%, specifying the response of the $\alpha$-th neuron to the current Cartesian, positional stimulus $P_{i}$.
The factor $\smash{(r^{p})^{D^{p}}}$ scales the neural activities, dependent on the relative distance $r^{p}$ between neighboring neurons. 
The individual neuron centers are evenly distributed in the expected range of the submodal stimuli.
The relative distance $r^{p}$ is also used to determine the diagonal variance entries:\\
\begin{equation}
\sigma^{p} = \zeta^{p} \cdot (r^{p})^2,
\end{equation}
where $\zeta^{p} \in (0, 1]$ modulates the breadth of the cell tunings. 
%  and denotes the continuity factor which ensures that for a stimuli in the expected range the sum of activity in the population is around 1.\\
Similarly, topological neurons' activations for direction and magnitude sub-modalities are computed by:
\begin{equation}
\dot{d}_{\alpha,i}(t)=(r^{d})^{D^{d}} \cdot N\left(d_{i}(t); c_{\alpha}^{d}, \Sigma^{d}\right),
\end{equation}
\begin{equation}
\dot{m}_{\alpha,i}(t)=(r^{m})^{D^{m}} \cdot N\left(m_{i}(t); c_{\alpha}^{m}, \Sigma^{m}\right),
\end{equation}
where the centers are set up to be in accordance with the dimension, range, and configuration space of the respective submodal stimuli.
Consequently, neurons that encode visual positional features ($D^{p}= 3$; $2$ for pendulum) are arranged evenly on a grid in a specific range.
Neurons that encode visual directional motion are arranged on the surface of a unit sphere ($D^{d}$ = 3; 2 for pendulum), while neurons that encode visual motion magnitudes are distributed linearly ($D^{m}$ = 1). 
Resulting population encoding activities are shown exemplarily in Figure~\ref{fig1:model}.

\subsection{Gestalt Perception and Feature Binding}
The binding problem concerns the selection and integration of separate visual features in the correct combination \cite{treisman1998feature}.
An approach for solving this problem is to selectively route respective feature positions and motion dynamics
such that the rerouted feature patterns match expected Gestalt dynamics.
% The model assumes that each bodily feature is processed by a specific neural processing path, such that features observed in an arbitrary order have to be put into the correct order.

Both the selection of features relevant for the recognition of biological motion and the assignment to the respective neural processing pathways are handled by an adaptive, gated connectivity matrix, which routes observed features $i\in \left \{ 1 ... N\right \}$ to bodily features $j\in \left \{ 1 ... M\right \}$ as follows:\\
\begin{equation}
\ddot{p_j}(t) = \sum_{i=1}^{N}w_{ij}(t) \cdot \dot{p_i}(t),
\end{equation}
\begin{equation}
\ddot{d_j}(t) = \sum_{i=1}^{N}w_{ij}(t) \cdot \dot{d_i}(t),
\end{equation}
\begin{equation}
\ddot{m_j}(t) = \sum_{i=1}^{N}w_{ij}(t) \cdot \dot{m_i}(t),
\end{equation}
where $\ddot{p_j}$, $\smash{\ddot{d_j}}$, and $\ddot{m_j}$ represent the population encoded activations of the j-th
assigned or bodily submodal feature in the position, motion direction, and motion magnitude domains, respectively. $\dot{p_j}$, $\smash{\dot{d_j}}$, and $\dot{m_j}$ represent the according activity of the i-th unassigned or observed submodal feature, and $w_{ij} \in (0, 1)$ represents the corresponding assignment strength.
The assignment strength is implemented by a non-linear neuron with logistic activation function:
\begin{equation}
w_{ij} = \frac{1}{1+exp(-w_{ij}^b)}
\end{equation}
where $w_{ij}^b$ denotes the activity of an adaptive parametric bias neuron.
Each set of submodal bodily feature populations is joined into a submodal Gestalt vector $g^x$:.
%That is, separate Gestalt vectors for posture, motion direction, and motion magnitude submodalities. The features within each set
%are concatenated in a following specified order:\\
\begin{equation}
g^{p}(t) =  (\ddot{p}_{1}(t), \ddot{p}_{2}(t), ..., \ddot{p}_{M}(t)),
\end{equation}
\begin{equation}
g^{d}(t) =  (\ddot{d}_{1}(t), \ddot{d}_{2}(t), ..., \ddot{d}_{M}(t)),
\end{equation}
\begin{equation}
g^{m}(t) =  (\ddot{m}_{1}(t), \ddot{m}_{2}(t), ..., \ddot{m}_{M}(t)),
\end{equation}
yielding Gestalt vectors for the posture, motion direction, and motion magnitude submodalities, respectively.

To learn distributed, predictive encodings of actions, each submodal Gestalt
perception $x\in\{p,d,m\}$ is encoded by one variational autoencoder \cite[VAE,][]{kingma2013auto}.
% that
%will implicitly extract the most descriptive components of the data when compressing it, and it is therefore a commonly used component in creative architectures. The idea behind VAE is relatively
%simple; push raw data (X) through a shrinking pipeline (q) and make the autoencoder network learn what features to remove and what to keep. Push the shrunken data (z) through an increasing sized pipeline (p), and make the autoencoder network learn what features it needs to add to reproduce the raw input, the empolyed VAE structure is also shown in the figure \ref{fig1:model}. The decoder is a generative model projecting from latent random variables to the input space. The encoder learns what kind of noise has to be inserted in the code layer, in order to make the decoder a good generative model.The reparametrization trick makes it possible to use backpropagation for training the whole model.
The bottom-up activated submodal Gestalt vectors $g^x$ are thus passed through the autoencoder, generating reconstructions of the Gestalt perceptions. 
As a consequence, when a fully trained autoencoder is provided with an imperfect stimulus (e.g. an observed action with unknown identity of the features, shown from an unknown perspective), it will tend to infer the closest known stimulus pattern, which will correspond to its (simulated) embodied, or self-perceptual experience.
The respective autoencoders thus generate posture, motion direction, and magnitude predictions. 
After model learning, the difference between the Gestalten input to the autoencoder and the regenerated Gestalten output is to be minimized by adapting the parametric bias neurons' activities of the binding matrix and the perspective taking modules. 
We denote the respective squared losses by $\mathcal{L}(p)$, $\mathcal{L}(d)$, and $\mathcal{L}(m)$, respectively. 
We scale the loss signals by respective factors $\beta^p$, $\beta^d$, and $\beta^m$ to balance the error signal influences. 
% are adapted with respect to the weighted error signals that originate from all submodal bodily features, as also indicated in Figure~\ref{fig1:model}.\\
% The positional error signals are represented by  represents error signal for motion magnitude features, which are again weighted by factors 
The actual adaptation of the parametric bias neurons' activities $w_{ij}^b(t)$ is computed by typical gradient descent with momentum: 
\begin{equation}
%\begin{split}
\Delta w_{ij}^b(t) = ~  -  \eta^f \frac{\partial \mathcal{L}(t)}{\partial w_{ij}^b(t)} + \gamma^f (w_{ij}^b(t-1) - w_{ij}^b(t-2)) , \label{eq:bindingudpate}
%\end{split}
\end{equation}
where $\gamma^f$ denotes the momentum, $\eta^f$ the learning rate for the feature binding adaptation process, and the loss signal $\mathcal{L}(t)$ equals to: 
\begin{equation}
\mathcal{L}(t) = \beta^p \mathcal{L}_p(t) + \beta^d \mathcal{L}_d(t) + \beta^m \mathcal{L}_m(t)
\end{equation}
During training, the assignment biases $w_{ij}^b$ are fixed to $w_{ii} = 1000$ for all $i$ (resulting in $w_{ii} \approx 1$) and to $w_{ij} = -1000$ for all $i\neq j$ (resulting in $w_{ij}\approx 0$), because the assignment is fixed during simulated self-observations.
During testing, all assignment biases are initialized to $-1$, resulting in an initial subtle mixture of all possible assignments, and are adapted over time by means of Eq~\ref{eq:bindingudpate}.

\subsection{Perspective Taking}
Perspective taking consists of a translation followed by a rotation of all considered visual features.
The employed mechanism is based on \cite{Schrodt:2015}.
It aims at establishing the best possible correspondence between the input and top-down expectations.
The translation determines the origin of the model’s internal, imagined frame of reference as well as the center of rotation. 
Perspective taking can be thought of as a mental transformation process that aligns the observer's perspective with a canonical, self-centered perspective. 
% allocentric perspectives with egocentric perspectives to make inferences based on embodied codes and self-experience.\\
The translation is determined by the bias neurons $b_a$, which, again, are adapted by gradient descent with momentum to minimize the top-down loss signal $\mathcal{L}(t)$:
\begin{equation}\label{eq18}
%\begin{split}
\Delta b_{a}(t) =~ -  \eta^b \frac{\partial \mathcal{L}(t)}{\partial^t b_{a}(t)}+  \gamma^b (b_{a}(t-1) - b_{a}(t-2))%\\
%\end{split}
\end{equation}
where $a \in \left \{x,y,z  \right \}$ denotes the affected axis, $\gamma^b$ the momentum term, and $\eta^b$ the adaptation rate. 
The translation biases $b_a$ are initialized to 0. 
Note that motion direction and magnitude are invariant to translations. 
As a result, the adaptation is determined the posture-respective weighted error signals $\beta^p \Delta_{1...M}^{p}$ only (cf. also Figure~\ref{fig1:model}).

Rotation is performed via a neural $3\times 3$ matrix $R$, which is driven by three Euler angles $\alpha_x$, $\alpha_y$ , and $\alpha_z$ each of which represents rotation around a specific Cartesian axis.
The corresponding pre-synaptic connection structure is shown in the following equation:
\begin{equation}
R = R_x(\alpha_x(t)) R_y(\alpha_y(t)) R_z(\alpha_z(t))
\end{equation}
Similar to the translation, the rotation is represented by bias neurons, which can be adapted online by gradient descent.
The adaptation over time follows the rule
\begin{equation}\label{eq21}
%\begin{split}
\Delta \alpha_{a}(t) =~  -  \eta^r \frac{\mathcal{L}(t)}{\partial^r \alpha_{a}(t)}+  \gamma^r (\alpha_{a}(t-1) - \alpha_{a}(t-2))\\
%\end{split}
\end{equation}
where $a \in \left \{x,y,z  \right \}$, $\gamma^r$ specifies the momentum, and $\eta^r$ the adaptation rate. 
Note that motion magnitude is invariant to rotation by nature and thus not considered in this process.
%\begin{equation}
%E^r = \left \{ \beta^p \Delta_{1...M}^{p}, \beta^d \Delta_{1...M}^{d} \right \}
%\end{equation}
The rotation biases are initialized to 0.
% Both, perspective taking and feature binding are driven by minimizing the differences between predicted sensory inputs governed by embodied encodings and observed sensory inputs.

\subsection{Related Work}

% MVB: We will integrate this part in our related work section below: 
%      which is a further development of research previously conducted by  “Anonymous (2018)” but with a standard generative autoencoder and the ability of switching to other generic motion patterns like 2 dimensional pendulum; 
% Therefore, when motion patterns of another person are observed, the model has to infer the frame of reference the action is currently observed from to establish the visual correspondence to the learned embodied codes \cite{schrodt2014modeling}. \\
% MVB: This is an implementational detail that is (unfortunately) not relevant for publication per se: 
%      furthermore our architecture has the potential to remove population coding and advance to the next layer with observed features alone which gives us the ability to investigate the effectiveness of using population coding.

From the brain-inspired side, dynamic neural fields are principally closely related to our model \cite{erlhagen2002dynamic,erlhagen2006dynamic} in that the activities in multiple dynamic neural fields strive for an overall consistent encoding. 
For example, they dynamically develop mappings between spatial task parameters that can resolve redundancies while generating motor plans \cite{erlhagen2002dynamic,martin2019process, martin2009redundancy}. 
More generally, it has been recently shown that dynamic neural fields can establish temporal bindings between different conceptual representations such as spatial arrangements of objects and compatible linguistic descriptions thereof \cite{Sabinasz:2020,schoner2019dynamics}.
Our perspective taking mechanisms can also be loosely related to neurocomputational models, which approximate actual representation formats identified in the brain \cite{Deneve:2004,pouget2000information,Pouget:2003}.

From the deep learning side, transformer networks \cite{Jaderberg:2015} are rather closely related in that an affine transformation is proposed, which works similar to our perspective taking module. 
In contrast to our work, though, the transformer network infers the frame of reference purely stimulus driven in a feed-forward manner. 
Our model couples the perspective taking module with a generative model, enabling retrospective inference of the perspective taking parameters. 
Moreover, various attentional mechanisms can be considered related \cite{Vaswani:2017} seeing that these architectures selectively process information. 
In contrast, our binding matrix flexibly reroutes information in the attempt to bind the information retrospectively rather than in a feed-forward manner, which may be more closely related to capsule networks \cite{Sabour:2017}. 
Our binding matrix adaptation mechanism may be most closely related to \citet{Memisevic:2013}, where gated autoencoder (or restricted Boltzmann machine) structures are learned. 
In our case, we apply the binding matrix, which is similar to gating, before the actual autoencoding.

\section{Experimental Results}
%\subsection{Used data}
We evaluate our model on a simple two-joint pendulum task as well as on more challenging 3D bodily motion capture data. 
For the latter, we use the  Motion Capture (MoCap) database of the Graphics Lab of the Carnegie Mellon University \cite{CMU}.
The CMU motion tracking data was recorded with $12$ high-resolution infrared cameras, each of which was recording at $120$Hz using $41$ tracking markers taped on jumpsuits of human subjects.
Every marker provided a 3D bodily landmark position, which was then mapped onto a skeleton file, which specifies limb connectivities and lengths.
Here we focus on continuous, cyclic actions and therefore use the motion capture data specified in Table~\ref{table_data}, which offers cyclic 3D walking motion patterns of three different participants. 
Out of $30$ limbs provided by the skeleton file, $M =$ 15  were selected as visual inputs, as shown in 
% where each was encoded by a three-dimensional Cartesian coordinate; also shown in 
Figure~\ref{TRainCompare} left.
To generate clean training data , we extracted a short episode of a walking trial of 260 time steps of Subject 35 and manually edited it to form a continuous cycle with 1036 frames in total.
We edited walking trials of Participants 5 and 6 in a similar manner, yielding 1000 frames in total, each.

% Furthermore, although CMU database provides many different actions, there are few, comparable and cyclic action examples that are performed by multiple actors.
%
\begin{table}[b!]
    \caption{Used data for training and testing the model}
    \label{table_data}
    \scriptsize
    \begin{tabularx}{\linewidth}{Xp{2.8cm}p{2.8cm}}
        \toprule
        & Training & Testing\\ \midrule
        3D Walking & Subject 35,\newline 7 cycles, 1036 frames \vspace{0.1cm} & Subjects 5 and 6, \newline ~7 cycles, 1000 frames \vspace{0.1cm}\\
        
        2D Pendulum & 6 cycles, 1000 frames & same pattern \\ \bottomrule
    \end{tabularx}
\end{table}

As another generic case we trained the model with generated two jointed pendulum data, which we adapted from mathplotlib's animation example \cite{Pendulum}.
It consist of 0.8 and 0.6 meter length plus 1.25 and 1 kilogram mass of the corresponding joints 1 and 2.
The data used for training and testing at the pendulum swing back and forth seven times with completely loose lower limb.
%$\sim$

If not stated differently, all results reported below show averages (and standard deviations where possible) over ten independently trained networks. 

%\begin{figure}[t]
%\centering
%\includegraphics{Mocap_Walk_Human.pdf}
%\caption{Shows a walking human subject; out of all possible joints green joints were selected as they seem to convey enough information to the model}
%\label{walker}
%\end{figure}

% We do not need to restate this: 
% Note that the visual inputs can potentially be transformed entirely via a translation vector $b^{data}$ and a rotation matrix $A^{data}$. Additionaly, to test the feature binding capability of the network the visual features can be provided in arbitrary order.

\subsection{VAE Prediction Error }
\begin{figure}[ht]
%\centering
\includegraphics[width=0.98\columnwidth,height=6.5cm]{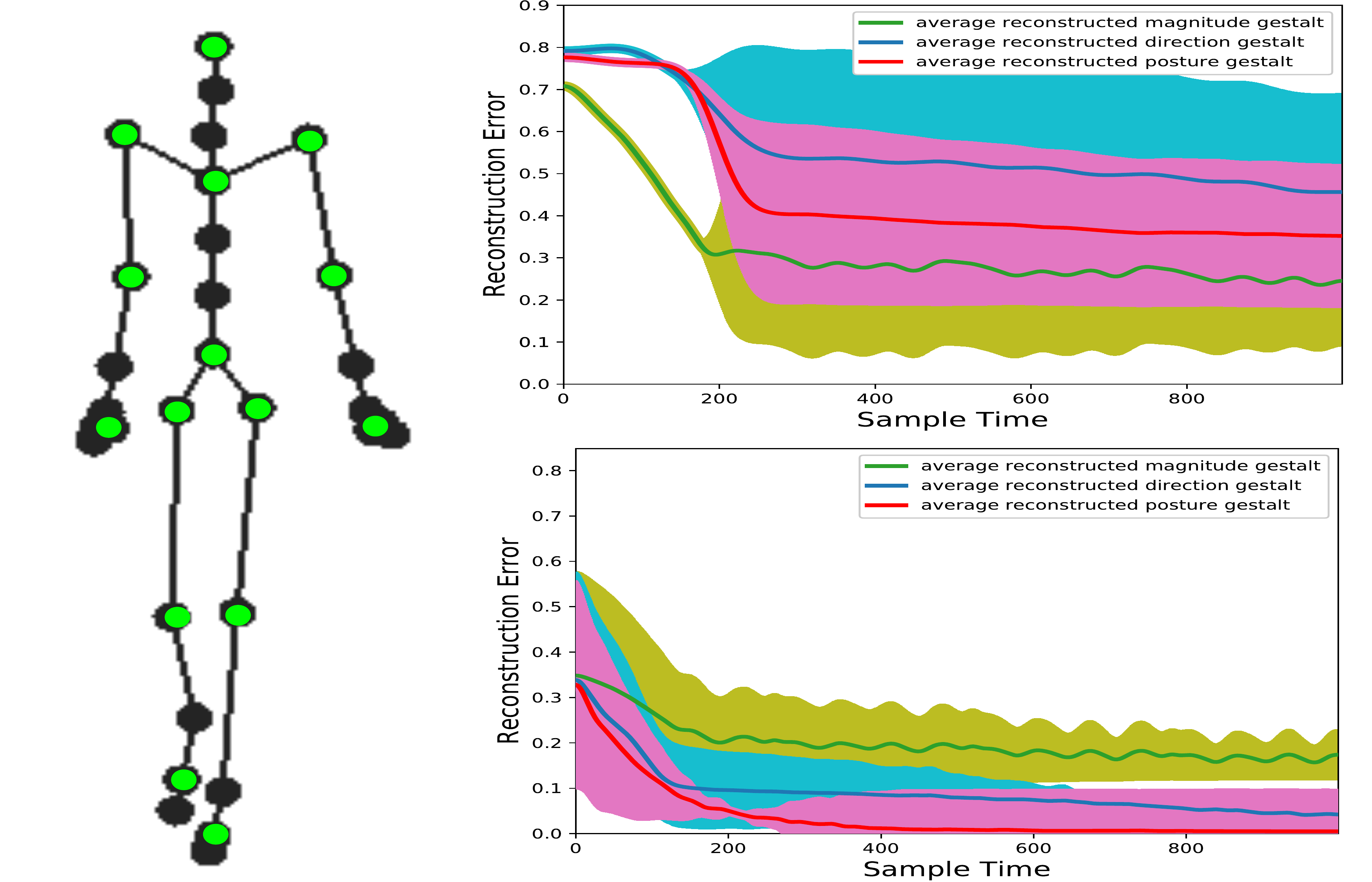} 
\caption{\textbf{left}: Motion capture skeleton data indicating the chosen 15 features; \textbf{right}: Visual spatial reconstruction error (i.e. Binary Cross-Entropy Loss) of posture, direction, and magnitude Gestalt perceptions; \textbf{right-up}: trained when using no population coding; \textbf{right-down}:trained using population coding}
\label{TRainCompare}
\end{figure}

During training the model has full access to visual information and visual stimuli are perceived from an egocentric point of view (perspective taking and feature binding adaptations are disabled). 
The aim is to analyze the compression quality of the autoencoder. 
We used the PyTorch library of Python. 
The chosen network parameters are specified in Table~\ref{train_config}.
Grid search was used to determine good parameter settings. 
While a more detailed analysis of parameter influences is beyond the scope of this paper, we can state that medium parameter value variations did not appear to change the reported results in any fundamental manner. 
To evaluate the influence of population coding on the model, we also trained another VAE model without the population encoding, feeding the raw Cartesian data into the VAE. 
In both cases, the model was able to significantly improve its reconstruction error over time, as can be clearly seen in Figure~\ref{TRainCompare} right.
Albeit the different error measures without and with population encoding cannot be directly compared, it is well-noticeable that learning takes place in both cases. 
Moreover, the relative improvement particularly of posture and motion direction Gestalt reconstruction error is much stronger in the case of population encoding (note the different ranges of the y-axes).

\begin{table*}[t!]
	\caption{Used data for training the variational autoencoder}
	\label{train_config}
	\scriptsize
	\begin{tabularx}{\linewidth}{p{3cm}XXXXXXXXX}
		\toprule
		Experiment & Learning Rate Pos & Learning Rate Dir & Learning Rate Mag & Optimization & Hidden Size & Latent Size & $\zeta^{p}$ & $\zeta^{d}$ & $\zeta^{m}$\\ \midrule
		Using population coding & $1\cdot10^{-3}$ & $8\cdot10^{-4}$ & $5\cdot10^{-4}$ & Adam & 45 & 25 & 0.85 & 0.85 & 0.95 \\ 
		Using raw data & $1\cdot10^{-3}$ & $2\cdot10^{-5}$ & $8\cdot10^{-4}$ & Adam & 25 & 10 & - & - & - \\
		Two Jointed Pendulum & $1\cdot10^{-2}$ & $1\cdot10^{-2}$ & $1\cdot10^{-3}$ & Adam & 45 & 25 & 0.85 & 0.85 & 0.95 \\ \bottomrule
	\end{tabularx}
\end{table*}

%\begin{figure}[htb]
%	\begin{minipage}[b]{0.27\columnwidth}
%		\centering
%		\includegraphics[width=1.1\textwidth]{figures/Mocap_Walk_Human.pdf}
%		\caption{Motion capture ske\-le\-ton data indica\-ting the chosen 15 fea\-tures.}
%		\label{walker}
%	\end{minipage}%
%\hspace{0.02\columnwidth}
%	\begin{minipage}[b]{0.71\columnwidth}
%		\centering
%		\includegraphics[width=0.99\textwidth,height=3.12cm]{figures/NoPopCodeTrainingAllRuns.pdf}
%		\includegraphics[width=0.99\textwidth,height=3.12cm]{figures/TrainingAllRuns.pdf}
%		\caption{Visual spatial reconstruction error (mean squared error / cross-entropy loss) of posture, direction, and magnitude Gestalt VAE reconstructions. Upper panel: no population coding; Lower panel: with population coding}
%		\label{TRainCompare}
%	\end{minipage}
%\end{figure}

%\begin{figure}[h]       
%    \includegraphics[width=0.27\columnwidth, height=6.5cm]{figures/Mocap_Walk_Human.pdf}
%    \hspace{10px}
%    \includegraphics[width=0.71\columnwidth, height=3.12cm]{figures/NoPopCodeTrainingAllRuns.pdf}
%    \hspace{10px}
%    \includegraphics[width=0.71\columnwidth, height=3.12cm]{figures/TrainingAllRuns.pdf}
%    \caption{this is the caption}
%    \label{materialflowChart}
%\end{figure}
%

%\begin{figure}[ht]
%\centering
%\includegraphics[width=0.9\columnwidth]{NPVAEtraining.pdf}
%\includegraphics[width=0.9\columnwidth]{PVAEtraining.pdf}
%\caption{Visual spatial reconstruction error (i.e. Binary Cross-Entropy Loss) of posture, direction, and magnitude Gestalt perceptions; \textbf{up}: trained when using no population coding; \textbf{down}:trained using population coding}
%\label{TRainCompare}
%\end{figure}

\subsection{Adaptation of Feature Binding}
While the results above show that the VAE does learn good encodings, the critical question in this work was whether the resulting signal is useful to accomplish feature binding and perspective taking, which we evaluate separately. 
%Here we evaluate feature binding and percepcapabilities are evaluated separatedly from the adaptive perspective taking components. 
To focus on feature binding, we disabled perspective taking.
Neural feature binding bias activities $w_{ij}^b$ were reset to $-5$ for each test trial, resulting in assignment strengths of $w_{ij}\approx 0.0067$ effectively distributing all feature value information uniformly over the VAE inputs. 
Please note that it is not necessary to permute the order of the inputs for the evaluation, since the model loses its knowledge about the correct assignment at this point.
The hyperparameters used during the feature binding experiments are shown in Table~\ref{table1}.
% The initial assignment strength furthermore decides on the initial speed of feature binding, since it marginally activates the codes by mixtures of all possible feature constellations, producing rather chaotic initial error signals with magnitudes proportional to the prior assignment strength.\\
% As also shown in the parameters, and as indicated before, the submodal error signals which are minimized for feature binding are individually weighted by the parameters $\beta^{pos}$, $\beta^{dir}$ and $\beta^{mag}$.\\

\begin{table}[t]
	\caption{Hyperparameters used for feature binding}
	\label{table1}
	\scriptsize
	\begin{tabularx}{\linewidth}{*{5}{p{1.67cm}}}
		\toprule
		Experiment 1 &
		Experiment 2 &
		Experiment 3 &
		Experiment 4 \\ \midrule
		No Pop Code &
		With Pop Code &
		With Pop Code &
		2D Pendulum\\
		$\beta^{pos}$= 5 &
		$\beta^{pos}$= 6 &
		$\beta^{pos}$= 8 &
		$\beta^{pos}$= 1 \\
		$\beta^{dir}$= 1 &
		$\beta^{dir}$= 0 &
		$\beta^{dir}$= 2 & 
		$\beta^{dir}$= 8 \\
		$\beta^{mag}$= 0.125 &
		$\beta^{mag}$= 0 &
		$\beta^{mag}$= 0.125 &
		$\beta^{mag}$= 2\\
		$\eta^f$= 1 &
		$\eta^f$= 1 &
		$\eta^f$= 1 &
		$\eta^f$= 0.1 &
		\\
		$\gamma^f$= 0.9 &
		$\gamma^f$= 0.9 &
		$\gamma^f$= 0.9 & 
		$\gamma^f$= 0.9 \\
		\bottomrule
	\end{tabularx}
\end{table}

\begin{figure}[t!]
	\hfill \includegraphics[width=0.5\textwidth]{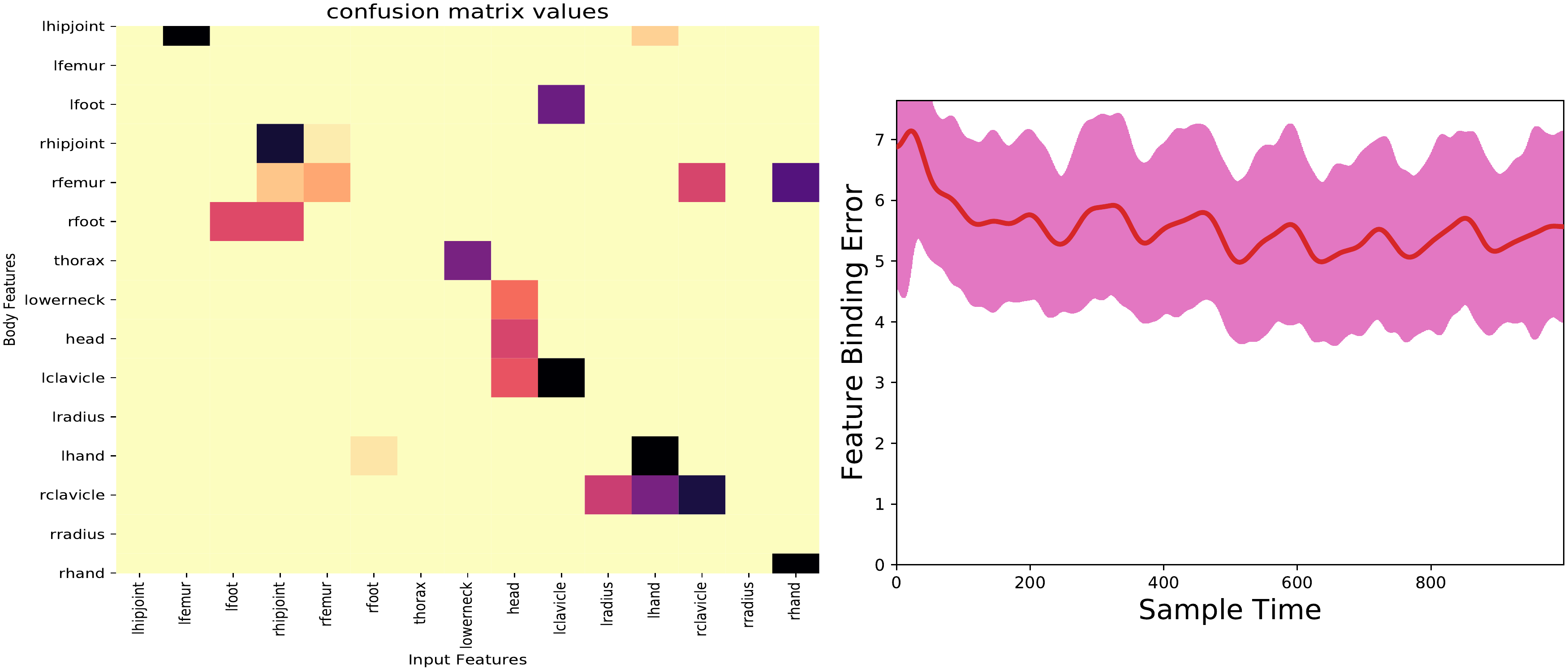}
	\hfill \includegraphics[width=0.5\textwidth]{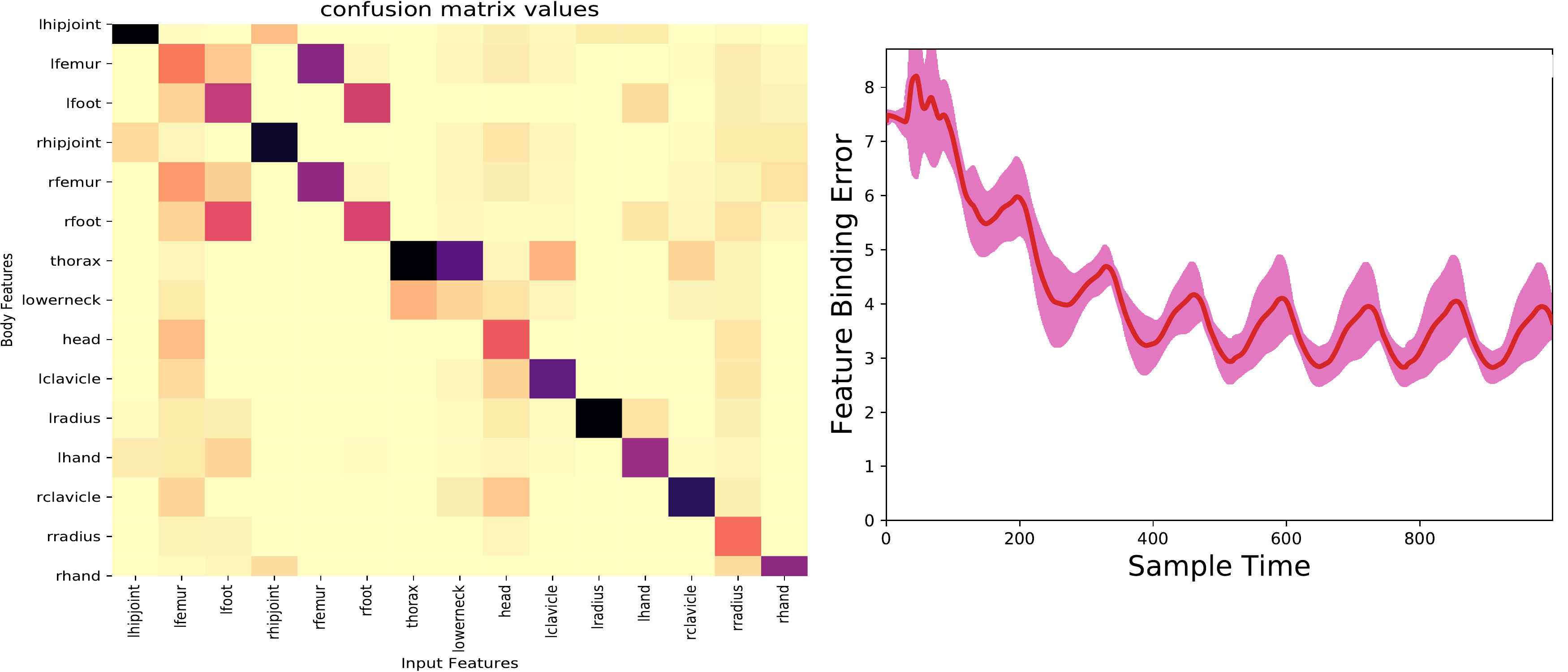}
	\hfill \includegraphics[width=0.5\textwidth]{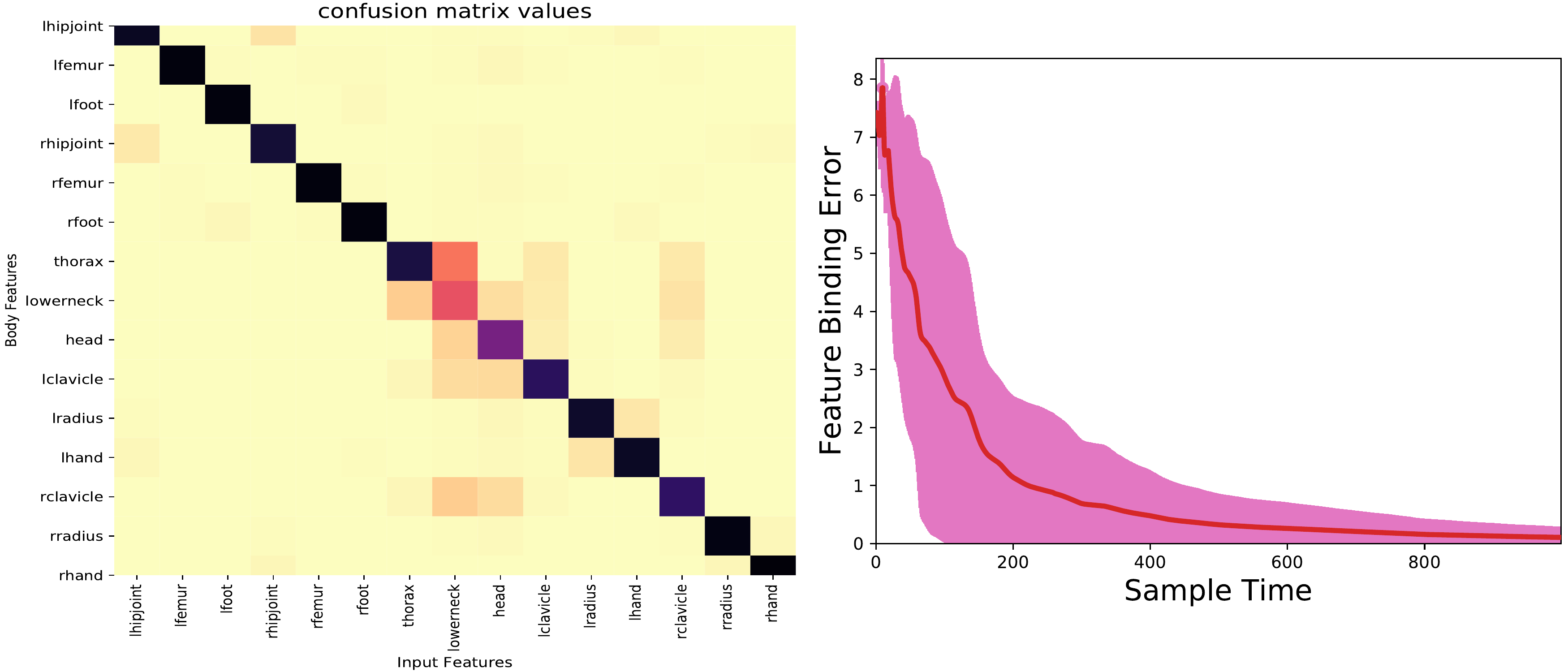}
	\hfill \includegraphics[width=0.25\textwidth]{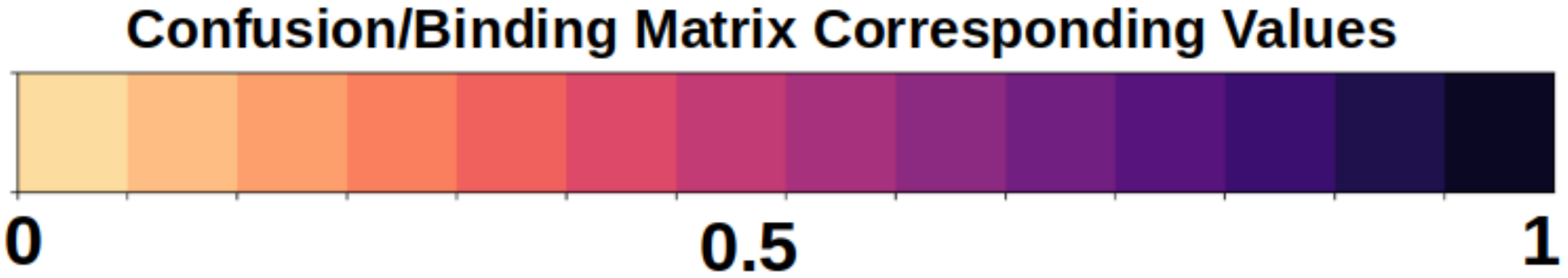} 
	\caption{Adapting the feature binding at first three experiments mentioned in Table~\ref{table1} }
	\label{3DFB}
\end{figure}

\begin{figure}[t!]
	\centering
	\includegraphics[width=0.99\columnwidth]{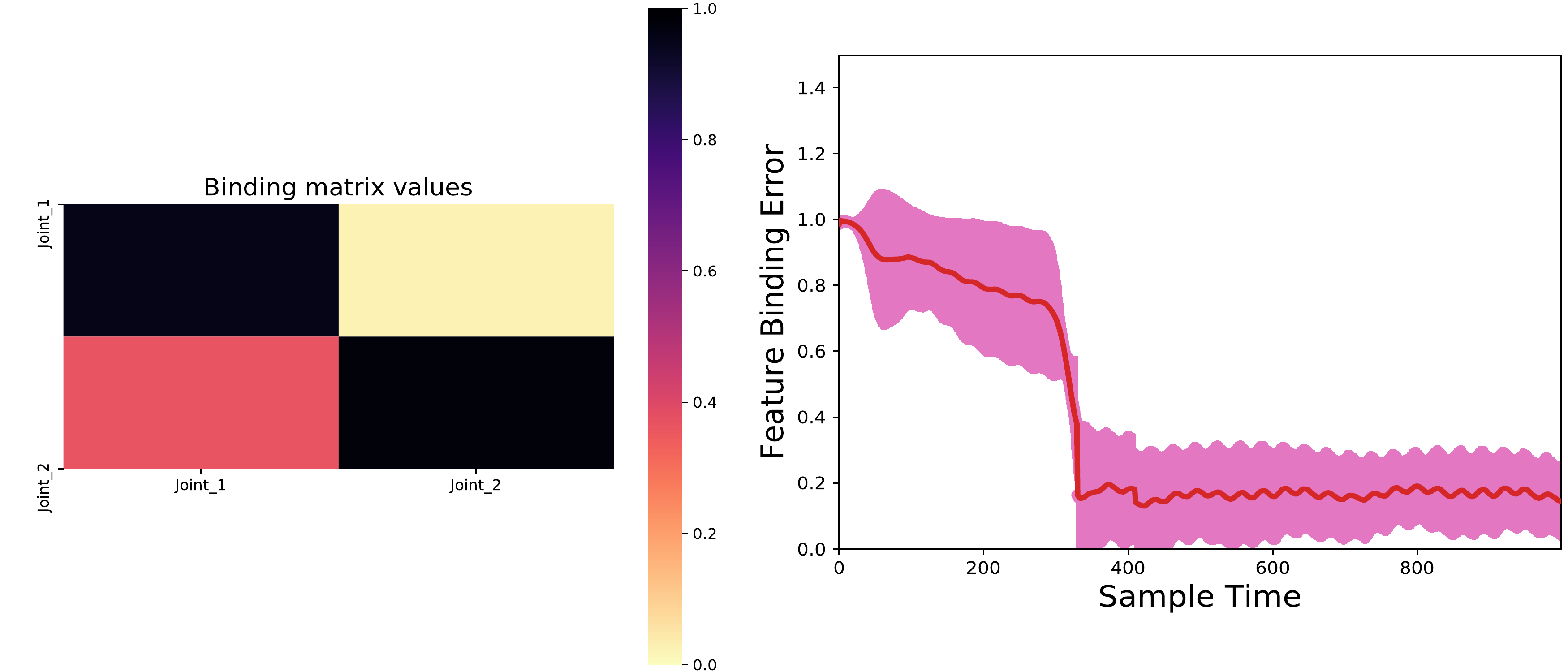} 
	\caption{Adapting the feature binding on two jointed pendulum}
	\label{2DFB}
\end{figure}

To measure the progress of feature binding in these experiments, we define a Feature Binding Error (FBE) as the sum of Euclidean distances between the model’s assignment of a bodily feature and the correct assignment:
\begin{equation}
 FBE(t) =\sum_{j=1}^{M}\sqrt{(w_{jj}(t)-1)^2 + \sum_{i=1,i\neq j}^{N}w_{ii}(t)^2 }
\end{equation}
Figure~\ref{2DFB} shows that the feature binding error decreases in all encoding cases. 
Moreover, Figure~\ref{3DFB} shows the resulting binding matrix values after 1000 steps of feature binding adaptation. 
Clearly, the applied population encoding is highly useful to decrease the FBE and thus to identify the right feature bindings, which corresponds to the diagonal. 
Please note again that the diagonal is identified without prior assumptions, that is, any other feature shuffling will yield the same result. 
It is interesting to analyze the binding matrix values, which essentially can be interpreted as a confusion matrix, in further detail. 
Without population encoding, full binding matrix convergence does not take place. 
For example, the hip joint is confused with the femur, and even severe confusions can be observed, such as the left foot with the left clavicle. 

With population encoding, the FBE drops significantly lower.
When only the posture encoding is considered, however, left-right and neighboring limb confusions remain. 
Only when motion information is added in population encoded form, hardly any confusions remain, indicating full Gestalt perception. 
It may be noted that the probability to guess the correct assignment by chance is virtually impossible (225 choose 15 yields a value of $9.1 \cdot 10^{22}$ possibilities). 
% in a single shot would be $1.3.10^{12}$ in this scenario, a-> not sure how this was calculated... if we calculate 225 choose 15 one gets a value of 9.1e22
The results clearly show that the VAE is able to identify the right bindings, particularly when population encodings are used and complementary feature information is provided. 
It is noticeable that the population encodings essentially enable the encoding of multi-modal distributions, which may be highly useful to test multiple binding options concurrently, most effectively guiding the binding values to the most consistent overall assignment over time. 
Additionally, the complementary information from posture, motion direction, and motion magnitude help to resolve remaining ambiguities.

%Our results show that instead of testing multiple assignments in parallel, nd that the model cannot independently keep up and test assignments in parallel, but only iteratively optimizes them using the described heuristics, using parallel and also conflicting population activity. Again, the model has no information about the derived error measures FBE. Nonetheless, these errors very steadily and robustly decrease in the progress of feature binding by means of gradient descent on the submodal reconstruction errors of the autoencoders, although the submodal errors do not continuously decrease.
%Thus, in a nutshell, feature binding continuously improves the inferred perception in the autoencoders. The inferred perception activates the submodal code and in turn generates the submodal expectation as driving signal for feature
%binding. 

To verify the generality of our results, we also evaluated binding performance on the pendulum data.
Figure~\ref{2DFB} confirms that binding also works robustly for the two joint pendulum data. 
Even if some latent activity of Joint 2 is added to Joint 1, this disruption seems to be minor, also indicating robust Gestalt perception.

%\begin{figure}[ht]
%\centering
%\includegraphics[width=0.35\columnwidth]{2DFBCMatrix.pdf} 
%\includegraphics[width=0.64\columnwidth]{2dFBerror.pdf}
%\caption{Adapting the feature binding on two jointed pendulum}
%\label{2DFB}
%\end{figure}

\subsection{Adaptation of Perspective Taking}
Focusing on perspective taking next, we fix the binding matrix to the correct diagonal binding values and assess the complexity when facing different perspectives onto a known motion pattern. 
Thus, we confront the fully trained models with trials of the test set, fix the binding matrix, and evaluate the development of spatial translation and orientation parametric biases. 
Each motion capture trial is first transformed by a random, three-dimensional, constant rotation offset, followed by a translation offset, before feeding it into the the model.
As a result, the model perceives the data from an unknown viewpoint and needs to transfer it into its known, egocentric frame of reference. 
% Thus, in other terms, the model perceives the data from an unknown, allocentric view-point, and is to transfer it into its known, egocentric frame of reference. \cite{schrodt2014modeling}.
As a quantitative measure for the transformation progress, we define an orientation difference (OD) measure: 
\begin{equation}
OD(t) = \frac{180}{2\Pi}Acos(\frac{tr(A^{model}(t)A^{data}(t))-1}{2}) \text{in}\,^{\circ},
\end{equation}
where $A^{data}$ is a constant rotation matrix applied to all visual
inputs and $A^{model}$ is the dynamic, currently inferred rotation matrix of the model. 
For measuring the translation with respect to the learned egocentric view, a translation difference (TD) measure is used:
\begin{equation}
TD(t)=\left \| b^{data}(t)-b^{model}(t) \right \|  \text{in cm},
\end{equation}
where $b^{data}$ is the constant per trial offset applied to the data and $b^{model}$ is the momentary adaptation of the model.
Employed model parameters are shown in Table~\ref{tablePT}.

\begin{table}[t]
	\caption{Hyperparameters used for perspective taking}
	\label{tablePT}
	\scriptsize
	\begin{tabularx}{\linewidth}{p{1.5cm} XX XX XX X}
		%	\begin{tabularx}{\linewidth}{p{3.8cm}|XX} %{|*{2}{p{4cm}|}}
	\toprule
	\textbf{Parameter} & $\eta^r$ & $\gamma^r$ & $\eta^b$ & $\gamma^b$ & $\beta^{pos}$ &$\beta^{dir}$ & $\beta^{mag}$  \\\midrule
	 \textbf{Value}	& $1.10^{-2}$ &  $9.10^{-1}$ & $8.10^{-2}$ & $9.10^{-1}$ & 8 & 3 & 0.125 \\ \bottomrule
\end{tabularx}
\end{table}

\begin{figure}[t!]
\includegraphics[width=0.9\columnwidth]{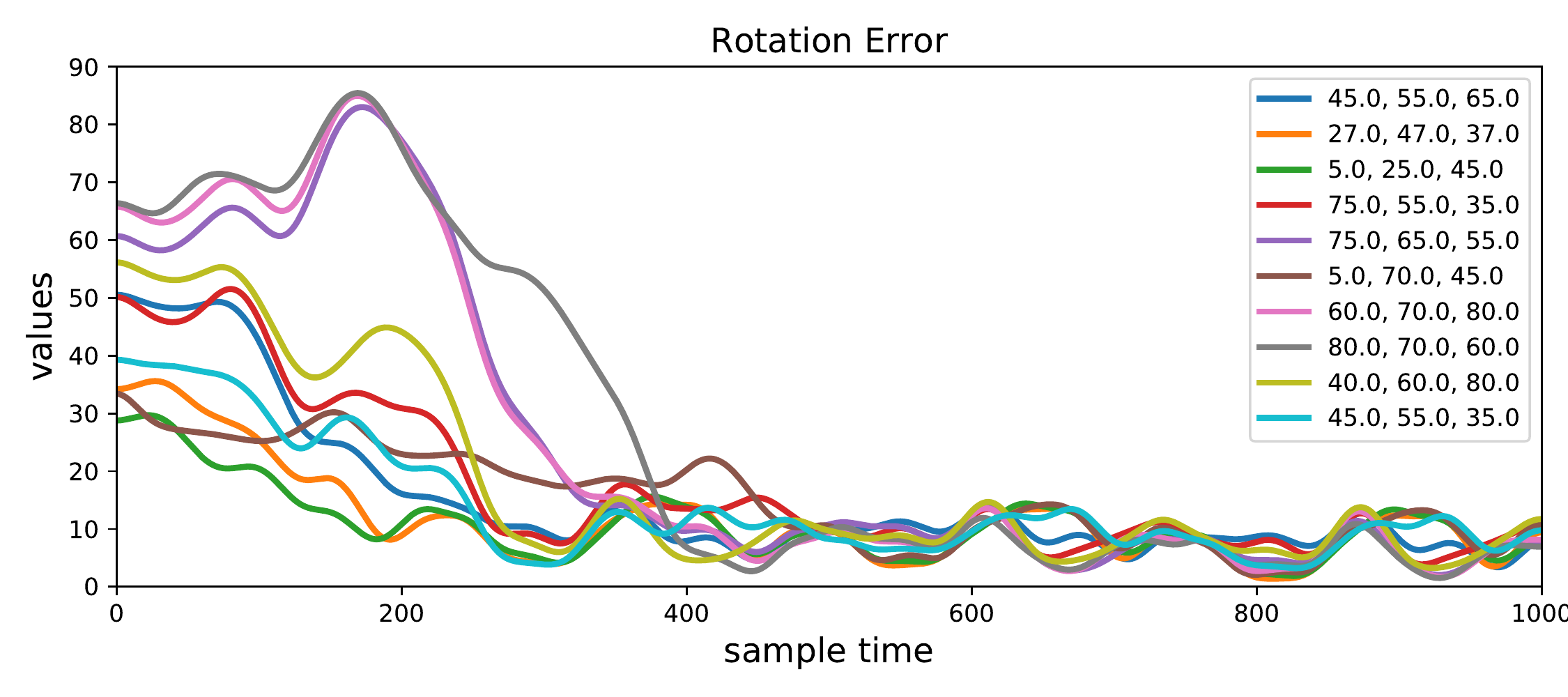}
\caption{Adapting the Perspective Taking: Rotation}
\label{RotaFig}
\includegraphics[width=0.9\columnwidth]{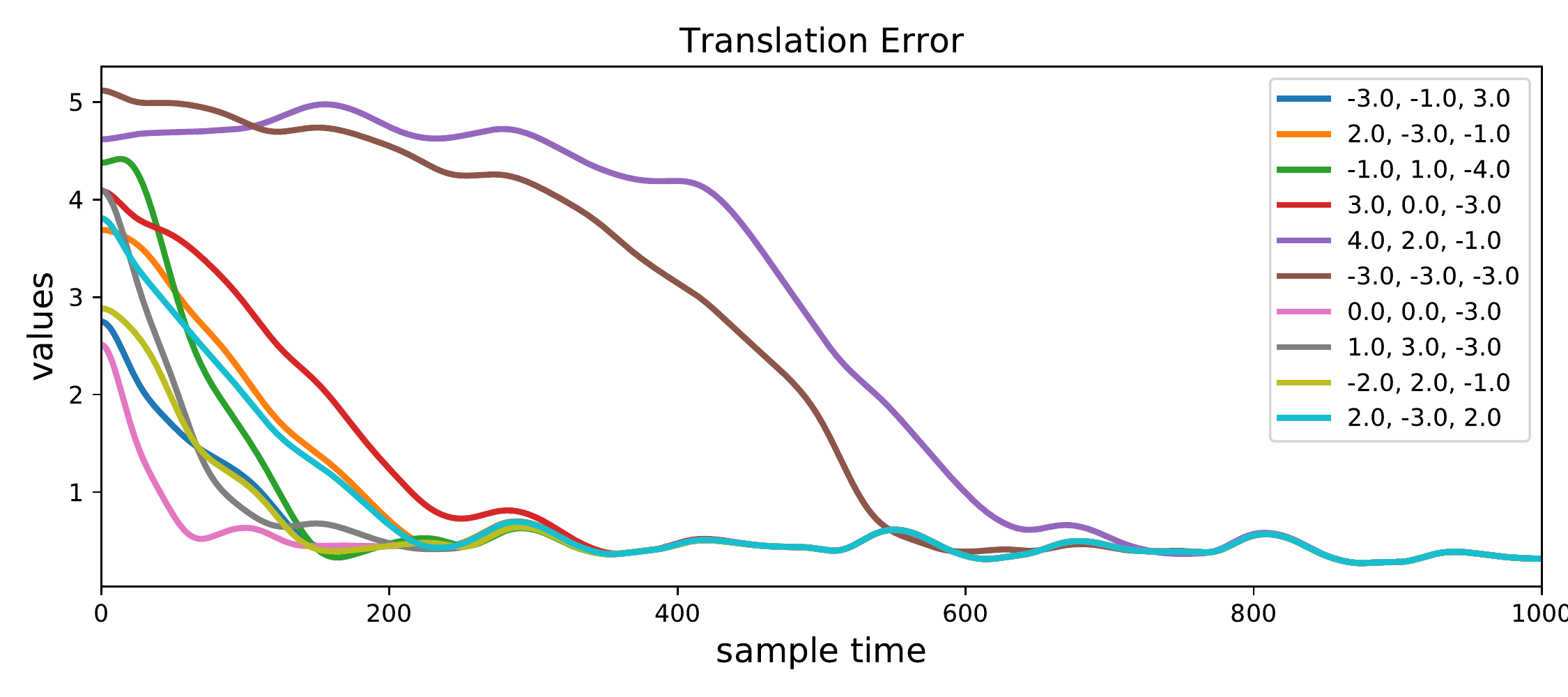}
\caption{Adapting the Perspective Taking: Translation}
\label{PTFig}
\end{figure}

Figure~\ref{RotaFig} shows the trained Rotation matrix while all visual input data was rotated with randomly assigned $\alpha_x$, $\alpha_y$ and $\alpha_z$ for one of the trained networks (results were qualitatively equivalent for the other trained networks).
Figure~\ref{PTFig} shows the corresponding translation bias inference during testing for the same network.
As can be seen, the stronger the perspective is disturbed, the longer it takes to infer the correct perspective. 
Rather extreme rotations of close to $90^\circ$ on all three axes are hardest. 
Similarly, strong translations yield delayed convergence, as can be expected.

\begin{figure}[htb!]
\includegraphics[width=1.0599\columnwidth]{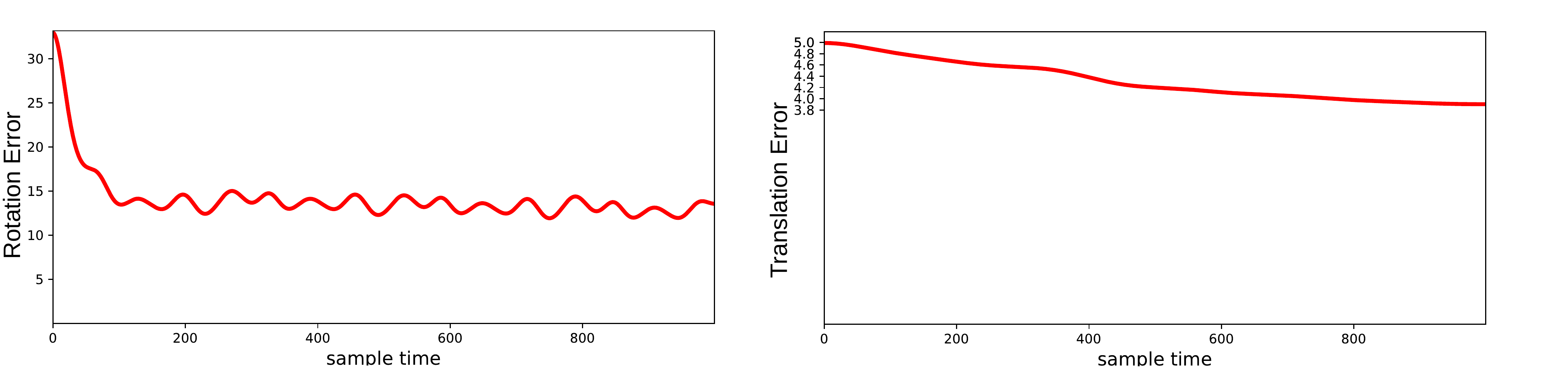}
\includegraphics[width=1.0599\columnwidth]{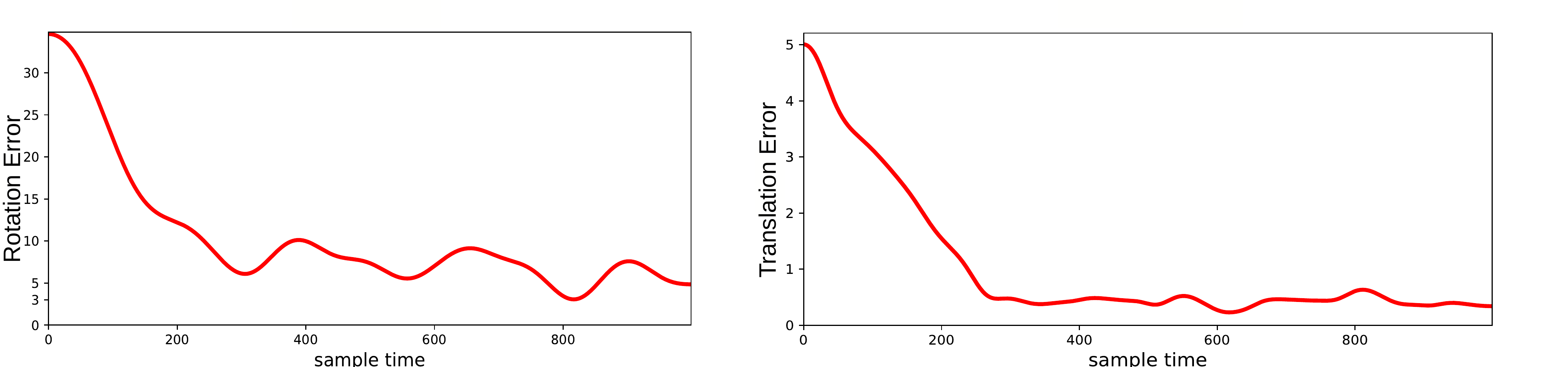}
\caption{Adapting rotation and translation; Upper-Row: no population coding; Lower-Row: With population coding}
\label{PTComparisions}
\end{figure}

To explore the perspective taking abilities further, we exemplarily look at the performance of a case where both rotations and translations are applied to the input data (we set $b_{x}= -2$ , $b_{y}= 2.5$, $b_{z}= -4$, $R_{x}= 25^\circ$ , $R_{y}= 35^\circ$, $R_{z}= 45^\circ$).
Figure~\ref{PTComparisions} shows that also in the case of perspective taking the population encoding is highly useful. 
Moreover, it confirms that translation and rotation distortions can be optimized concurrently. 

%\begin{table}[htb!]
%    \caption{Exempardistortion of origin and orientation of perspective}
%    \label{table_Compare}
%    \scriptsize
%    \begin{tabularx}{\linewidth}{XX}
%        \toprule
%        Experiment 1 &
%        Experiment 2 \\ \midrule
%        No Population Coding &
%        With Population Coding \\
%        $b_{x}$= -2 , $b_{y}$= 2.5, $b_{z}$= -4 &
%        $b_{x}$= -2 , $b_{y}$= 2.5, $b_{z}$= -4\\
%        $R_{x}$= 25 , $R_{y}$= 35, $R_{z}$= 45&
%        $R_{x}$= 25 , $R_{y}$= 35, $R_{z}$= 45 \\
% \bottomrule
%    \end{tabularx}
%\end{table}

\section{Summary and Conclusion}
We have introduce a generative neural network architecture, which tackles the perspective taking and binding challenges. 
In contrast to related approaches, such as attention-based architectures \cite{Vaswani:2017} or transformer networks \cite{Jaderberg:2015}, our system applies retrospective inference tuning parametric bias neurons \cite{Sugita:2011a,Tani:2004,Tani:2017}, which are in our case dedicated to establish feature bindings and adapt the internal perspective onto the observed features. 
The flexible binding is most closely related to tripartite autoencoder and Restricted Boltzmann machine approaches \cite{Memisevic:2013} as well as to capsule networks \cite{Sabour:2017}. 
In contrast, our model applies a more direct, retrospective gradient-based adaptation mechanism to latent, dedicated parametric bias neurons, adhering to the principle of endowing neural network architectures with effective inductive learning and inference biases \cite{Battaglia:2018}.
As a result, starting from a canonical perspective on a biological motion pattern, we have shown that perspective taking and binding works highly reliably, particularly when complementary spatio-temporal feature patterns are available and encoded with population codes. 
Interestingly, the resulting ability can be related to embodied social cognition, mirror neurons, perspective taking, and Gestalt perception.

Despite the success in both feature binding and perspective taking online, future work is necessary to investigate scalability and robustness of our model. 
In order to be able to distinguish multiple Gestalt patterns, another top-down module needs to be added, which may be able to identify distinct motion patterns, such a walking, running, jumping, but also static patterns, such as sitting, standing, or lying down (cf. \citealp{Butz:2019,Schrodt:2016}; \cite{Schrodt:2018}. 
Moreover, we intend to extend the model with an LSTM-based temporal encoder-decoder architecture \cite{Bahdanau:2015}, in order to reap additional information from the temporal dynamics retrospectively \cite{Butz:2019}.
Seeing that we did not need to add any regularization methods, residual connections, or other techniques useful for scaling-up deep ANN \cite{Goodfellow:2016}, we are confident that our approach will be applicable to larger dataset and disjunct Gestalt patterns.

Overall, we hope that our technique will be useful also in other domains, where information needs to be flexibly bound together and associated to other data on the fly, striving for overall consistency.
Seeing that perspective taking and binding problems are universal problems in cognitive science, the proposed architecture and retrospective inference mechanisms may be very useful also in solving related challenges in other domains besides biological motion patterns.

\vfill

\clearpage
\newpage

\bibliography{RNNBib}

\begin{thebibliography}{46}
\providecommand{\natexlab}[1]{#1}
\providecommand{\url}[1]{\texttt{#1}}
\providecommand{\urlprefix}{URL }
\expandafter\ifx\csname urlstyle\endcsname\relax
  \providecommand{\doi}[1]{doi:\discretionary{}{}{}#1}\else
  \providecommand{\doi}{doi:\discretionary{}{}{}\begingroup
  \urlstyle{rm}\Url}\fi

\bibitem[{Pen(2017)}]{Pendulum}
 2017.
\newblock Mathplotlib animation Examples; double pendulum Release: 2.0.2.
\newblock \url{https://matplotlib.org/examples/animation/}.

\bibitem[{CMU(2018)}]{CMU}
 2018.
\newblock Carnegie Mellon University Motion Capture Database.
\newblock \url{http://mocap.cs.cmu.edu}.

\bibitem[{Bahdanau, Cho, and Bengio(2015)}]{Bahdanau:2015}
Bahdanau, D.; Cho, K.; and Bengio, Y. 2015.
\newblock Neural Machine Translation by Jointly Learning to Align and
  Translate.
\newblock \emph{International Conference on Learning Representations} .

\bibitem[{Barsalou(1999)}]{Barsalou:1999}
Barsalou, L.~W. 1999.
\newblock Perceptual symbol systems.
\newblock \emph{Behavioral and Brain Sciences} 22: 577--600.

\bibitem[{Battaglia et~al.(2018)Battaglia, Hamrick, Bapst, Sanchez-Gonzalez,
  Zambaldi, Malinowski, Tacchetti, Raposo, Santoro, Faulkner, Gulcehre, Song,
  Ballard, Gilmer, Dahl, Vaswani, Allen, Nash, Langston, Dyer, Heess, Wierstra,
  Kohli, Botvinick, Vinyals, Li, and Pascanu}]{Battaglia:2018}
Battaglia, P.~W.; Hamrick, J.~B.; Bapst, V.; Sanchez-Gonzalez, A.; Zambaldi,
  V.; Malinowski, M.; Tacchetti, A.; Raposo, D.; Santoro, A.; Faulkner, R.;
  Gulcehre, C.; Song, F.; Ballard, A.; Gilmer, J.; Dahl, G.; Vaswani, A.;
  Allen, K.; Nash, C.; Langston, V.; Dyer, C.; Heess, N.; Wierstra, D.; Kohli,
  P.; Botvinick, M.; Vinyals, O.; Li, Y.; and Pascanu, R. 2018.
\newblock Relational inductive biases, deep learning, and graph networks.
\newblock \emph{arXiv preprint 1806.01261} .

\bibitem[{Buschman and Miller(2007)}]{buschman2007top}
Buschman, T.~J.; and Miller, E.~K. 2007.
\newblock Top-down versus bottom-up control of attention in the prefrontal and
  posterior parietal cortices.
\newblock \emph{Science} 315(5820): 1860--1862.

\bibitem[{Butz et~al.(2019)Butz, Bilkey, Humaidan, Knott, and Otte}]{Butz:2019}
Butz, M.~V.; Bilkey, D.; Humaidan, D.; Knott, A.; and Otte, S. 2019.
\newblock Learning, planning, and control in a monolithic neural event
  inference architecture.
\newblock \emph{Neural Networks} 117: 135--144.
\newblock \doi{10.1016/j.neunet.2019.05.001}.

\bibitem[{Butz and Kutter(2017)}]{Butz:2017}
Butz, M.~V.; and Kutter, E.~F. 2017.
\newblock \emph{How the Mind Comes Into Being: Introducing Cognitive Science
  from a Functional and Computational Perspective}.
\newblock Oxford, UK: Oxford University Press.

\bibitem[{Castiello et~al.(2002)Castiello, Lusher, Mari, Edwards, and
  Humphreys}]{castiello2002observing}
Castiello, U.; Lusher, D.; Mari, M.; Edwards, M.; and Humphreys, G. 2002.
\newblock Observing a human or a robotic hand grasping an object: Differential
  motor priming effects.
\newblock In Prinz, W.; and Hommel, B., eds., \emph{Common Mechanisms in
  Perception and Action: Attention and Performance}. Oxford University Press.

\bibitem[{Cook et~al.(2014)Cook, Bird, Catmur, Press, and Heyes}]{Cook:2014}
Cook, R.; Bird, G.; Catmur, C.; Press, C.; and Heyes, C. 2014.
\newblock Mirror neurons: From origin to function.
\newblock \emph{Behavioral and Brain Sciences} 37: 177--192.
\newblock \doi{10.1017/S0140525X13000903}.

\bibitem[{Den\`eve and Pouget(2004)}]{Deneve:2004}
Den\`eve, S.; and Pouget, A. 2004.
\newblock Bayesian multisensory integration and cross-modal spatial links.
\newblock \emph{Journal of Physiology - Paris} 98: 249--258.

\bibitem[{Edwards, Humphreys, and Castiello(2003)}]{edwards2003motor}
Edwards, M.~G.; Humphreys, G.~W.; and Castiello, U. 2003.
\newblock Motor facilitation following action observation: A behavioural study
  in prehensile action.
\newblock \emph{Brain and Cognition} 53(3): 495--502.

\bibitem[{Erlhagen and Bicho(2006)}]{erlhagen2006dynamic}
Erlhagen, W.; and Bicho, E. 2006.
\newblock The dynamic neural field approach to cognitive robotics.
\newblock \emph{Journal of neural engineering} 3(3): R36.

\bibitem[{Erlhagen and Sch{\"o}ner(2002)}]{erlhagen2002dynamic}
Erlhagen, W.; and Sch{\"o}ner, G. 2002.
\newblock Dynamic field theory of movement preparation.
\newblock \emph{Psychological review} 109(3): 545.

\bibitem[{Gallese and Goldman(1998)}]{Gallese:1998}
Gallese, V.; and Goldman, A. 1998.
\newblock Mirror neurons and the simulation theory of mind-reading.
\newblock \emph{Trends in Cognitive Sciences} 2(12): 493--501.

\bibitem[{Gallese, Keysers, and Rizzolatti(2004)}]{gallese2004unifying}
Gallese, V.; Keysers, C.; and Rizzolatti, G. 2004.
\newblock A unifying view of the basis of social cognition.
\newblock \emph{Trends in Cognitive Sciences} 8(9): 396--403.

\bibitem[{Goodfellow, Bengio, and Courville(2016)}]{Goodfellow:2016}
Goodfellow, I.; Bengio, Y.; and Courville, A. 2016.
\newblock \emph{Deep Learning}.
\newblock Cambridge, MA: MIT Press.

\bibitem[{Heyes(2001)}]{heyes2001causes}
Heyes, C. 2001.
\newblock Causes and consequences of imitation.
\newblock \emph{Trends in cognitive sciences} 5(6): 253--261.

\bibitem[{Jackson, Meltzoff, and Decety(2006)}]{jackson2006neural}
Jackson, P.~L.; Meltzoff, A.~N.; and Decety, J. 2006.
\newblock Neural circuits involved in imitation and perspective-taking.
\newblock \emph{Neuroimage} 31(1): 429--439.

\bibitem[{Jaderberg et~al.(2015)Jaderberg, Simonyan, Zisserman, and
  kavukcuoglu}]{Jaderberg:2015}
Jaderberg, M.; Simonyan, K.; Zisserman, A.; and kavukcuoglu, k. 2015.
\newblock Spatial Transformer Networks.
\newblock In Cortes, C.; Lawrence, N.~D.; Lee, D.~D.; Sugiyama, M.; and
  Garnett, R., eds., \emph{Advances in Neural Information Processing Systems
  28}, 2017--2025. Curran Associates, Inc.

\bibitem[{Johnson and Demiris(2005)}]{Johnson:2005}
Johnson, M.; and Demiris, Y. 2005.
\newblock Perceptual Perspective Taking and Action Recognition.
\newblock \emph{International Journal of Advanced Robotic Systems} 2(4):
  301--309.

\bibitem[{Jäkel et~al.(2016)Jäkel, Singh, Wichmann, and Herzog}]{Jaekel:2016}
Jäkel, F.; Singh, M.; Wichmann, F.~A.; and Herzog, M.~H. 2016.
\newblock An overview of quantitative approaches in Gestalt perception.
\newblock \emph{Quantitative Approaches in Gestalt PerceptionVision Research}
  126: 3--8.
\newblock \doi{10.1016/j.visres.2016.06.004}.

\bibitem[{Kingma and Welling(2013)}]{kingma2013auto}
Kingma, D.~P.; and Welling, M. 2013.
\newblock Auto-encoding variational bayes.
\newblock \emph{arXiv preprint arXiv:1312.6114} .

\bibitem[{Koffka(2013)}]{koffka2013principles}
Koffka, K. 2013.
\newblock \emph{Principles of Gestalt psychology}, volume~44.
\newblock Routledge.

\bibitem[{Martin, Reimann, and Sch{\"o}ner(2019)}]{martin2019process}
Martin, V.; Reimann, H.; and Sch{\"o}ner, G. 2019.
\newblock A process account of the uncontrolled manifold structure of joint
  space variance in pointing movements.
\newblock \emph{Biological Cybernetics} 113(3): 293--307.

\bibitem[{Martin, Scholz, and Sch{\"o}ner(2009)}]{martin2009redundancy}
Martin, V.; Scholz, J.~P.; and Sch{\"o}ner, G. 2009.
\newblock Redundancy, self-motion, and motor control.
\newblock \emph{Neural Computation} 21(5): 1371--1414.

\bibitem[{Meltzoff and Prinz(2002)}]{Meltzoff:2002}
Meltzoff, A.~N.; and Prinz, W. 2002.
\newblock \emph{The imitative mind: Development, evolution and brain bases}.
\newblock Cambridge: Cambridge University Press.

\bibitem[{Memisevic(2013)}]{Memisevic:2013}
Memisevic, R. 2013.
\newblock Learning to Relate Images.
\newblock \emph{Pattern Analysis and Machine Intelligence, IEEE Transactions
  on} 35(8): 1829--1846.
\newblock ISSN 0162-8828.
\newblock \doi{10.1109/TPAMI.2013.53}.

\bibitem[{Moll and Meltzoff(2011)}]{Moll:2011}
Moll, H.; and Meltzoff, A.~N. 2011.
\newblock Perspective-taking and its foundation in joint attention.
\newblock In \emph{Perception, Causation, and Objectivity}, 286--304.

\bibitem[{Nehaniv, Dautenhahn et~al.(2002)}]{nehaniv2002correspondence}
Nehaniv, C.~L.; Dautenhahn, K.; et~al. 2002.
\newblock The correspondence problem.
\newblock \emph{Imitation in animals and artifacts} 41.

\bibitem[{Olshausen and Field(1997)}]{Olshausen:1997}
Olshausen, B.~A.; and Field, D.~J. 1997.
\newblock Sparse coding with an overcomplete basis set: a strategy employed by
  V1?
\newblock \emph{Vision Res} 37: 3311--25.

\bibitem[{Pouget, Dayan, and Zemel(2000)}]{pouget2000information}
Pouget, A.; Dayan, P.; and Zemel, R. 2000.
\newblock Information processing with population codes.
\newblock \emph{Nature Reviews Neuroscience} 1(2): 125--132.

\bibitem[{Pouget, Dayan, and Zemel(2003)}]{Pouget:2003}
Pouget, A.; Dayan, P.; and Zemel, R.~S. 2003.
\newblock Inference and computation with population codes.
\newblock \emph{Annual Review of Neuroscience} 26: 381--410.

\bibitem[{Sabinasz et~al.(2020)Sabinasz, Richter, Lins, and
  Schöner}]{Sabinasz:2020}
Sabinasz, D.; Richter, M.; Lins, J.; and Schöner, G. 2020.
\newblock Speaker-specific adaptation to variable use of uncertainty
  expressions.
\newblock \emph{Proceedings of the 42nd Annual Meeting of the Cognitive Science
  Society} 620--627.

\bibitem[{Sabour, Frosst, and Hinton(2017)}]{Sabour:2017}
Sabour, S.; Frosst, N.; and Hinton, G.~E. 2017.
\newblock Dynamic Routing Between Capsules.
\newblock In Guyon, I.; Luxburg, U.~V.; Bengio, S.; Wallach, H.; Fergus, R.;
  Vishwanathan, S.; and Garnett, R., eds., \emph{Advances in Neural Information
  Processing Systems 30}, 3856--3866. Curran Associates, Inc.

\bibitem[{Sch{\"o}ner(2019)}]{schoner2019dynamics}
Sch{\"o}ner, G. 2019.
\newblock The dynamics of neural populations capture the laws of the mind.
\newblock \emph{Topics in Cognitive Science} \doi{10.1111/tops.12453}.

\bibitem[{Schrodt(2018)}]{Schrodt:2018}
Schrodt, F. 2018.
\newblock \emph{Neurocomputational Principles of Action Understanding:
  Perceptual Inference, Predictive Coding,and Embodied Simulation}.
\newblock Ph.D. thesis, Faculty of Science, University of Tübingen.
\newblock \doi{10.15496/publikation-24327}.

\bibitem[{Schrodt and Butz(2016)}]{Schrodt:2016}
Schrodt, F.; and Butz, M.~V. 2016.
\newblock Just Imagine! Learning to Emulate and Infer Actions with a Stochastic
  Generative Architecture.
\newblock \emph{Frontiers in Robotics and AI} \doi{10.3389/frobt.2016.00005}.

\bibitem[{Schrodt et~al.(2014)Schrodt, Layher, Neumann, and
  Butz}]{schrodt2014modeling}
Schrodt, F.; Layher, G.; Neumann, H.; and Butz, M.~V. 2014.
\newblock Modeling perspective-taking upon observation of 3D biological motion.
\newblock In \emph{4th International Conference on Development and Learning and
  on Epigenetic Robotics}, 305--310. IEEE.

\bibitem[{Schrodt et~al.(2015)Schrodt, Layher, Neumann, and
  Butz}]{Schrodt:2015}
Schrodt, F.; Layher, G.; Neumann, H.; and Butz, M.~V. 2015.
\newblock Embodied Learning of a Generative Neural Model for Biological Motion
  Perception and Inference.
\newblock \emph{Frontiers in Computational Neuroscience} 9(79).
\newblock \doi{10.3389/fncom.2015.00079}.

\bibitem[{Sugita, Tani, and Butz(2011)}]{Sugita:2011a}
Sugita, Y.; Tani, J.; and Butz, M.~V. 2011.
\newblock Simultaneously emerging Braitenberg codes and compositionality.
\newblock \emph{Adaptive Behavior} 19: 295--316.
\newblock \doi{10.1177/1059712311416871}.

\bibitem[{Tani(2017)}]{Tani:2017}
Tani, J. 2017.
\newblock \emph{Exploring Robotic Minds}.
\newblock Oxford, UK: Oxford University Press.

\bibitem[{Tani, Ito, and Sugita(2004)}]{Tani:2004}
Tani, J.; Ito, M.; and Sugita, Y. 2004.
\newblock Self-organization of distributedly represented multiple behavior
  schemata in a mirror system: reviews of robot experiments using RNNPB.
\newblock \emph{Neural Networks} 17(8-9): 1273--1289.
\newblock \doi{10.1016/j.neunet.2004.05.007}.

\bibitem[{Treisman(1998)}]{treisman1998feature}
Treisman, A. 1998.
\newblock Feature binding, attention and object perception.
\newblock \emph{Philosophical Transactions of the Royal Society of London.
  Series B: Biological Sciences} 353(1373): 1295--1306.

\bibitem[{Tversky and Hard(2009)}]{Tversky:2009}
Tversky, B.; and Hard, B.~M. 2009.
\newblock Embodied and disembodied cognition: Spatial perspective-taking.
\newblock \emph{Cognition} 110(1): 124 -- 129.
\newblock \doi{10.1016/j.cognition.2008.10.008}.

\bibitem[{Vaswani et~al.(2017)Vaswani, Shazeer, Parmar, Uszkoreit, Jones,
  Gomez, Kaiser, and Polosukhin}]{Vaswani:2017}
Vaswani, A.; Shazeer, N.; Parmar, N.; Uszkoreit, J.; Jones, L.; Gomez, A.~N.;
  Kaiser, L.~u.; and Polosukhin, I. 2017.
\newblock Attention is All you Need.
\newblock In Guyon, I.; Luxburg, U.~V.; Bengio, S.; Wallach, H.; Fergus, R.;
  Vishwanathan, S.; and Garnett, R., eds., \emph{Advances in Neural Information
  Processing Systems 30}, 5998--6008. Curran Associates, Inc.

\end{thebibliography}
\end{document}